\documentclass[conference]{IEEEtran}
\IEEEoverridecommandlockouts
\usepackage{booktabs}
\usepackage{threeparttable}
\usepackage{stfloats}
\usepackage{cite}
\usepackage{amsmath,amssymb,amsfonts}
\usepackage{algorithmic}
\usepackage{graphicx}
\usepackage{textcomp}
\usepackage{xcolor}
\usepackage{spconf}
\usepackage{array}

\usepackage{fancyhdr} 
 
\newcommand{\eat}[1]{}

\def\BibTeX{{\rm B\kern-.05em{\sc i\kern-.025em b}\kern-.08em
    T\kern-.1667em\lower.7ex\hbox{E}\kern-.125emX}}
    
\begin{document}

\title{TimeRAG: BOOSTING LLM Time Series Forecasting 
via \\ Retrieval-Augmented Generation
\\
}    

\name{ Silin Yang$^{1,}$\sthanks{Equal Contribution} \qquad Dong Wang$^{2 , \ast}$ \qquad Haoqi Zheng$^{2,}$\sthanks{Corresponding Authors} \qquad Ruochun Jin$^{2, \dagger}$   }
\address{$^{1}$School of Computer Science, Peking University, Beijing, China\\
        $^{2}$College of Computer, National University of Defense Technology, Changsha, China}

\maketitle

\begin{abstract}
Although the rise of large language models (LLMs) has introduced new opportunities for time series forecasting,
existing LLM-based solutions require excessive training and exhibit limited transferability.
In view of these challenges, we propose TimeRAG,
a framework that incorporates Retrieval-Augmented Generation (RAG) into time series forecasting LLMs,
which constructs a time series knowledge base from historical sequences,
retrieves reference sequences from the knowledge base that exhibit similar patterns to the query sequence measured by Dynamic Time Warping (DTW),
and combines these reference sequences and the prediction query as a textual prompt to the time series forecasting LLM.
Experiments on datasets from various domains show that the integration of RAG improved the prediction accuracy of the original model by 2.97\% on average.

\end{abstract}


\section{Introduction}

Time series forecasting is critical in data science and machine learning research,
covering wide applications including financial market analysis, demand forecasting, weather prediction, etc.
Although deep-model-based forecasting methods such as LSTM \cite{sagheer2019time},  Reformer\cite{kitaev2020reformer} and Informer \cite{zhou2021informer} 
have achieved satisfactory performance on classical benchmarks \cite{lim2021time},
they can hardly capture the hidden complex patterns and dependencies
in large-scale sequential data with staggering complexity and diversity \cite{ye2024survey}.
In view of this challenge, researchers have explored the possibility of applying large language models (LLMs)
to time series analysis and prediction across various domains \cite{xue2023promptcast,zhou2023one}, since LLMs have demonstrated
remarkable achievements in natural language processing.
However, existing time series forecasting LLMs cannot easily adapt to
different domains, as the training of LLMs is computationally costly and
typically optimized for a specific domain \cite{jiang2024empowering}.
Moreover, due to the ``hallucination'' of LLMs \cite{bang2023multitask},
LLMs may generate inaccurate predictions, outliers,
or fabricated patterns when performing time series forecasting, that do not align with the data, with no interpretability.

In order to resolve these issues, we propose to boost
time series forecasting LLMs via Retrieval-Augmented Generation (RAG) \cite{lewis_retrieval-augmented_2020}.
Specifically, we first establish a time series knowledge base by collecting representative
sequential data from the training set via K-means clustering.
Then given the time series forecasting query as input,
we employ Dynamic Time Warping (DTW) \cite{muller2007dynamic} as the distance metric
to retrieve sequences, that share similar waveforms and trends
with the query, from the time series knowledge base as referential sequences,
since DTW is tolerant to temporal distortions.
Finally, the input query and the referential sequences are rewritten as
a natural language prompt and fed into the LLMs for prediction.
We have experimentally verified our method on M4 datasets, a collection of varying-frequencies time series from different domains\cite{makridakis2020m4},
where significant improvements of up to \textbf{13.12\%} have been observed.

Different from existing time series LLMs that require massive training costs \cite{jin_time-llm_2024,ansari_chronos_2024}
and previous RAG solutions\cite{lewis_retrieval-augmented_2020,liu2022uni},
to the best of our knowledge,
we are the first to propose a RAG framework specifically designed for time series data prediction 
without modifying the foundational parameters of the underlying LLM. 
Experimental results confirm that our method exhibits strong competitiveness when compared to both similar LLMs and baseline models.


The key contributions of our work are as follows:
\begin{itemize}
\item To the best of our knowledge,
we are the first to boost time series forecasting LLMs by Retrieval-Augmented Generation,
which significantly improves prediction accuracy.
\item We employ K-means clustering and Dynamic Time Warping to efficiently construct
a time series knowledge base, which facilitates the LLM to easily
adapt to different domains of time series.
\item We experimentally verify that RAG contributes to an average improvement of \textbf{2.97\%} in the accuracy of sequence forecasting.
\end{itemize}

\begin{figure*}[htbp]
	\centering		\includegraphics[width=\textwidth ]{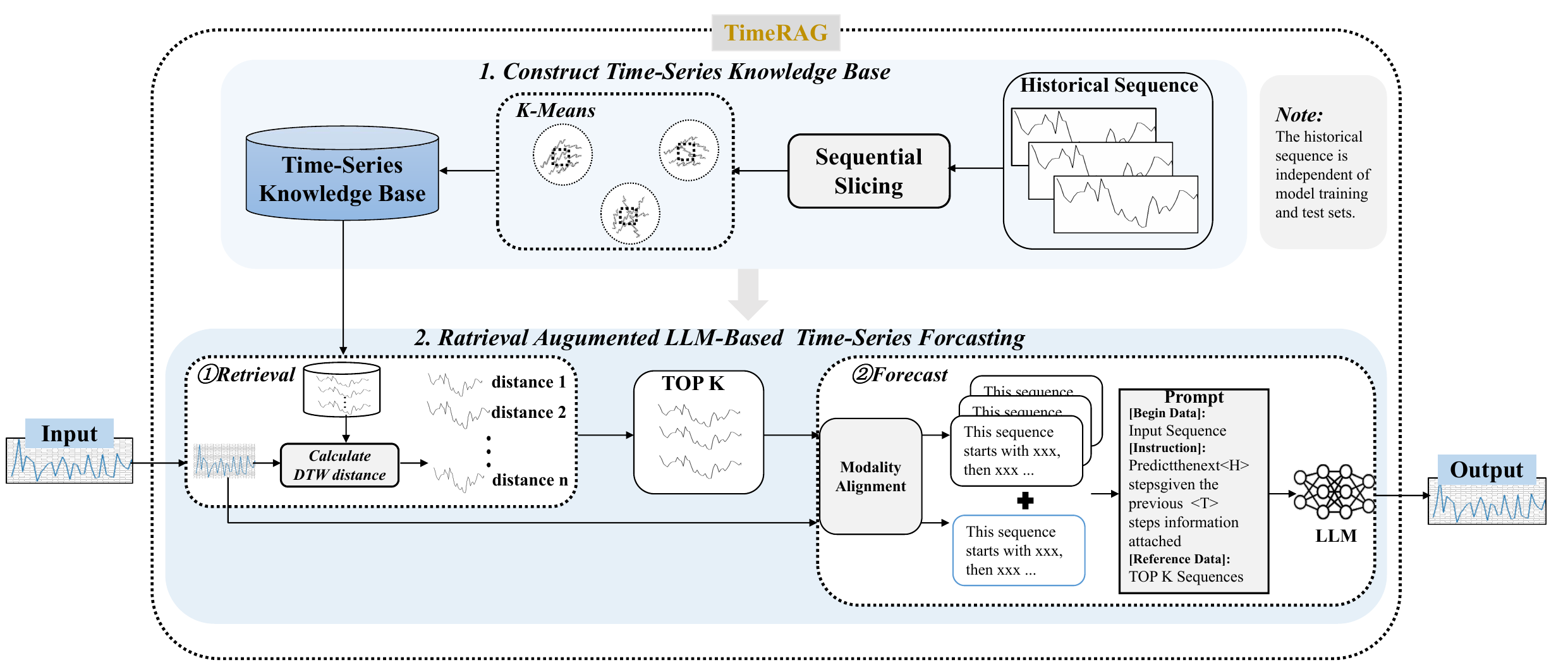}
	\centering
	\caption{Overview of our mechanism. }\label{fig overview}
\end{figure*}

\section{Method}
\subsection{Overview}
As shown in Fig. \ref{fig overview}, in order to enhance the performance of LLMs in time-series forecasting tasks,
we propose a retrieval-augmented framework, named TimeRAG, that
consists of two main components: a Time Series Knowledge Base (KB) (\ref{sec 3.2}) and an LLM-based time series forecasting model(\ref{sec 3.3}).
Specifically, TimeRAG first sequentially slices the original sequence into segments
and establishes a time series knowledge base by extracting representative segments from the training set using K-means clustering. 
Then input the time series forecasting query, we apply Dynamic Time Warping (DTW) as the distance metric to retrieve sequences from the time series knowledge base that exhibit similar waveforms and trends to the query, leveraging DTW’s ability to handle temporal distortions. The input query and the retrieved sequences are then reformulated into a natural language prompt, which is subsequently input into the LLMs for prediction.


\subsection{Time-Series Knowledge Base} \label{sec 3.2}
In order to establish the Time-Series Knowledge Base,
TimeRAG first performs sequential slicing on the original sequence through a sliding window,
and employs clustering to select representative segments for storage.
Instead of storing and retrieving complete raw sequences,
our sequence-segmentation approach preserves the local information of the sequence,
avoids long sequences where LLMs tend to miss key information\cite{liu-etal-2024-lost},
and improves the retrieval efficiency.
Specifically, given a sequence $X = (x_t, ... x_{t+n}), X \in R^n$ of time-varying values
from time $t$ to $t+n$,
TimeRAG first adopts a sliding window approach with a step size of $S$ and a window length of $L$
to slice $X$ into several sub-sequences $X_L$, where $X_L \in R^L$.
Secondly, K-means clustering is applied to these fragments for capturing representative sequences. 
Given a set of $N$ sequence fragments $Q_L = \{X_{L_i}\}, i \in [1,N]$,
K-means first initializes a set of cluster centroids $C = $  $ \{X_{c_1},..., X_{c_k} \}, X_{c_i} \in Q_L, i \in [1,k]$ and assign each $X_{L_i}$ to the closest centroid, 
where the distances between each sequence fragment $X_{L_i}$
and all centroids are measured by the Euclidean distance
as Eq.\ref{kmeans}:
\begin{equation}
d = \| X_{L_i} - X_{c_j} \|_2 \label{kmeans}
\end{equation}
where $X_{L_i} \neq X_{c_j}$.
After this initialization,
K-means iteratively updates each centroid as the mean of sequences within each cluster
and reassigns each sequence to the cluster whose centroid is the closest to the sequence,
which gradually minimizes the total sum of distances between all points and their corresponding cluster centroids. 
Finally, TimeRAG constructs the Time-series Knowledge Base by collecting the sequence that is the closest to its centroid from each cluster.




\subsection{Retrieval-Augmented LLM-based Time-series Forecasting} \label{sec 3.3}

Although LLMs have demonstrated remarkable performance in time series forecasting\cite{liu-etal-2024-lstprompt},
their prediction accuracy deteriorates when processing sequences that have not been previously trained. 
Moreover, LLMs show general performance degradation due to their tendency to forget\cite{liu-etal-2024-lost},
which may adversely affect the accuracy of time series forecasting.
In view of these challenges,
we introduce the retrieval-augmented LLM-based time-series forecasting
that consists of the following two stages:
(1) retrieval of similar sequences based on DTW\cite{sakoe1978dynamic},
and (2) prediction by LLM where both the original sequence and the retrieved similar sequences are combined to enhance forecast accuracy.


In the retrieval stage,  
given the prediction query and the Time-Series Knowledge Base,
TimeRAG employs DTW to retrieve top-$K$ sequences that are most similar to the query from the knowledge base. 
Specifically, given the input query sequence $X_{input},\ X_{input} \in R^n$ for prediction, 
DTW first constructs an $n \times L$ matrix for each sequence $X_L$ in the knowledge base, 
where the element $(i,j)$ of the matrix represents the distance $d(i,j)$ 
between the $X_{input_i}$ and  $X_{L_j}$, which represent the i-th point of $X_{input}$  and the j-th point of $X_{L}$ respectively. The distance $d(i,j)$ is computed as following formula:
\begin{equation}
d(i,j) = (X_{input_i},X_{L_j})^2 \label{dis}
\end{equation}
We refer to the path $W$ from matrix element $(1,1)$ to $(n, L)$, consisting of several adjacent and non-repeating matrix elements, as the warping path,
where the m-th element of $W$ is defined as $w_m = d(m_i,m_j)$, which is computed by Eq. \ref{dis}.
Thus, $W$ can be given by:
\begin{equation}
W = w_1,...w_m,...w_M \label{path}
\end{equation}
where $max(n,L) \leq M \leq n + L$  and $w_M = d(n,L)$.

 The algorithm then employs dynamic programming to obtain the shortest warping path, which can be utilized to measure the similarity between $X_{input}$ and $X_L$ as Eq.\ref{dtw}: 
\begin{equation}
Simi(X_{input}, X_L) = min \frac{\sqrt{\sum_{m=1}^{M}{w_m}}}{M} \label{dtw}
\end{equation}
where $Simi(X_{input}, X_L)$ \ denotes the similarity between the $X_{input}$ and $X_L$.
Finally, TimeRAG selects the top $K$ sequences that are most similar to the query sequence as the retrieval results,
measured by $Simi$.


In the model prediction stage, TimeRAG follows Time-LLM \cite{jin_time-llm_2024} that adopts a reprogramming layer to align the sequence modality with the natural language modality.
As shown in Fig. \ref{fig overview}, the input query sequence $X_{input}$ and the retrieved sequences are transformed through the reprogramming layer and concatenated as one prompt, which enhances the prediction performance of the LLM.



\section{Experiments}

\begin{table}[]
\centering
\tiny
\caption{The summary of the M4 dataset and the knowledge base.
The number of sequences in the knowledge base is larger than
the total quantity since one original sample can be sliced into multiple sequence segments.}
\resizebox{0.95\linewidth}{!}{
\begin{tabular}{@{}ccccc@{}}
\toprule
Time          & Input        & Prediction  & Total    & KB  \\ 
Interval      & Length       & Length      & Quantity & Size           \\ \midrule
Yearly        & 12           & 6           & 23000    & 57400   \\
Quarterly     & 16           & 8           & 24000    & 50000   \\
Monthly       & 36           & 18          & 48000    & 60000   \\
Weekly        & 26           & 13          & 359      & 895     \\
Daily         & 28           & 14          & 4227     & 10565   \\
Hourly        & 96           & 48          & 414      & 1035    \\
Total         &  -           & -           & 100000   & 179895  \\ \bottomrule
\end{tabular}
}
\label{tab1}
\end{table}

\subsection{Experiments Setup}
\textbf{Datasets}.
We evaluate TimeRAG on the M4 benchmark,
a widely used dataset for time series forecasting that contains data from diverse domains, 
including finance, demographics, marketing, etc.,
with different sequential sampling frequencies: yearly, quarterly, monthly, weekly, daily, and hourly. 
Each frequency corresponds to specific prediction horizons and input lengths,
which supports the comprehensive evaluation of forecasting models.
More details of the dataset are provided in Tab. \ref{tab1}.

\textbf{Evaluation Metric}.
As introduced in \cite{jin_time-llm_2024},
we adopt the following three widely accepted metrics for performance evaluation:
(1) Symmetric Mean Absolute Percentage Error (SMAPE):
as a widely recognized measure in time series forecasting,
SMAPE quantifies forecast accuracy relative to actuals by computing the percentage error. 
(2) Mean Absolute Scaled Error (MASE):
this metric evaluates a model's predictive accuracy relative to a naive forecast strategy,
offering scale independence and robustness across a series of varying magnitudes.
(3) Overall Weighted Average (OWA):
drawing from the methodology in N-BEATS \cite{oreshkin2019n},
OWA integrates SMAPE and MASE to provide a holistic assessment of model performance. 
The smaller values of prediction results measured by SMAPE, MASE, and OWA,
the higher prediction accuracy the model achieves.

\textbf{Baselines}.
We compare TimeRAG with state-of-the-art time series models, including Transformer-based methods: iTransformer\cite{liu2023itransformer}, FEDformer\cite{zhou2022fedformer}, Pyraformer\cite{liu2021pyraformer}, Autoformer\cite{chen2021autoformer}, Informer\cite{zhou2021informer}, and Reformer\cite{kitaev2020reformer};
as well as other competitive models: Time-LLM\cite{jin_time-llm_2024}, DLinear\cite{zeng2023transformers},  TSMixer\cite{ekambaram2023tsmixer}, MICN\cite{wang2023micn}, FiLM\cite{zhou2022film} and LightTS\cite{campos2023lightts}.

\textbf{Training Settings}.
Inspired by \cite{jin_time-llm_2024}, TimeRAG employs the reprogramming technique
where the input time series are reprogrammed with text prototypes
before fed into a frozen LLM to align the two modalities. 
In order to obtain a well-trained reprogramming layer,
we trained TimeRAG based on Llama3 with a maximum of 50 training epochs,
using 8 A100 GPUs, Adam optimizer, and SMAPE as the loss function.  
To mitigate over-fitting, we employ dynamic learning rate adjustment and an early stopping strategy,
with the maximum learning rate set as 0.01.

\textbf{Knowledge Base Implementation}.
As sequential data shows distinct waveform characteristics at different time frequencies, 
we built separate knowledge bases for each frequency in the M4 dataset and split the remaining data into training, validation, and test sets.
Tab. \ref{tab1} shows statistics of the knowledge bases for the M4 dataset at different frequencies.
Once the knowledge bases were constructed,
TimeRAG enhanced the test samples by retrieving the \textbf{top-five} most relevant entries,
measured by DTW, from the corresponding knowledge base for each test case,
which enables retrieval-augmented time series forecasting.


\begin{table*}[ht]
\centering
\LARGE
\renewcommand\arraystretch{1.5}
\caption{Forecasting results on M4 dataset. 
\underline{{\color[HTML]{0000FF} Blue}} means the model is top-three at the current frequency and metric.}
\resizebox{\textwidth}{!}{%
\begin{tabular}{@{}c|ccc|ccc|ccc|ccc|ccc|ccc|ccc@{}}
\toprule
\multicolumn{1}{c|}{Frequency} &
  \multicolumn{3}{c|}{Yearly} &
  \multicolumn{3}{c|}{Quarterly} &
  \multicolumn{3}{c|}{Monthly} &
  \multicolumn{3}{c|}{Weekly} &
  \multicolumn{3}{c|}{Daily} &
  \multicolumn{3}{c|}{Hourly} &
  \multicolumn{3}{c}{\textbf{Average}} \\ \midrule
\multicolumn{1}{c|}{Metric} &
  SMAPE &
  MASE &
  OWA &
  SMAPE &
  MASE &
  OWA &
  SMAPE &
  MASE &
  OWA &
  SMAPE &
  MASE &
  OWA &
  SMAPE &
  MASE &
  OWA &
  SMAPE &
  MASE &
  OWA &
  SMAPE &
  MASE &
  OWA \\ \midrule

Autoformer &
  20.900 &
  3.741 &
  1.138 &
  14.913 &
  1.745 &
  1.324 &
  18.023 &
  1.715 &
  1.331 &
  16.758 &
  4.445 &
  1.509 &
  4.851 &
  4.756 &
  1.469 &
  26.874 &
  8.847 &
  2.001 &
  17.053 &
  4.208 &
  1.462 \\
DLinear &
  15.913 &
  2.755 &
  0.873 &
  \underline{{\color[HTML]{0000FF} 10.817}} &
  \underline{{\color[HTML]{0000FF} 1.203}} &
  \underline{{\color[HTML]{0000FF} 0.949}} &
  \underline{{\color[HTML]{0000FF} 13.448}} &
  1.191 &
  0.957 &
  13.714 &
  4.000 &
  1.294 &
  4.065 &
  3.883 &
  1.215 &
  \underline{{\color[HTML]{0000FF} 17.085}} &
  \underline{{\color[HTML]{0000FF} 3.847}} &
  \underline{{\color[HTML]{0000FF} 1.081}} &
  \underline{{\color[HTML]{0000FF} 12.507}} &
  \underline{{\color[HTML]{0000FF} 2.813}} &
  1.062 \\
FEDformer &
  15.347 &
  2.703 &
  0.829 &
  11.223 &
  1.292 &
  0.989 &
  14.129 &
  1.273 &
  1.014 &
  \underline{{\color[HTML]{0000FF} 11.824}} &
  \underline{{\color[HTML]{0000FF} 3.367}} &
  \underline{{\color[HTML]{0000FF} 1.102}} &
  \underline{{\color[HTML]{0000FF} 3.442}} &
  \underline{{\color[HTML]{0000FF} 3.259}} &
  \underline{{\color[HTML]{0000FF} 1.024}} &
  18.504 &
  4.968 &
  1.217 &
  \underline{{\color[HTML]{0000FF} 12.412}} &
  \underline{{\color[HTML]{0000FF} 2.810}} &
  \underline{{\color[HTML]{0000FF} 1.029}} \\
FiLM &
  15.850 &
  2.724 &
  0.846 &
  11.054 &
  1.291 &
  0.981 &
  \underline{{\color[HTML]{0000FF} 13.436}} &
  1.190 &
  0.956 &
  \underline{{\color[HTML]{0000FF} 12.385}} &
  3.778 &
  \underline{{\color[HTML]{0000FF} 1.195}} &
  3.637 &
  3.536 &
  1.096 &
  22.328 &
  9.654 &
  1.985 &
  13.115 &
  3.696 &
  1.177 \\
Informer &
  16.717 &
  2.919 &
  0.899 &
  15.664 &
  1.911 &
  1.421 &
  18.750 &
  1.810 &
  1.395 &
  14.307 &
  4.191 &
  1.353 &
  5.066 &
  4.820 &
  1.511 &
  36.071 &
  21.378 &
  4.026 &
  17.763 &
  6.172 &
  1.768 \\
iTransformer &
  19.762 &
  3.506 &
  1.071 &
  11.659 &
  1.330 &
  1.022 &
  16.653 &
  1.547 &
  1.215 &
  15.860 &
  3.930 &
  1.383 &
  4.040 &
  3.763 &
  1.192 &
  19.195 &
  5.513 &
  1.316 &
  14.528 &
  3.265 &
  1.200 \\
LightTS &
  \underline{{\color[HTML]{0000FF} 15.238}}  &
  2.680 &
  \underline{{\color[HTML]{0000FF} 0.817}} &
  \underline{{\color[HTML]{0000FF} 10.868}} &
  \underline{{\color[HTML]{0000FF} 1.228}} &
  \underline{{\color[HTML]{0000FF} 0.948}} &
  \underline{{\color[HTML]{0000FF} 13.203}} &
  \underline{{\color[HTML]{0000FF} 1.148} }&
  \underline{{\color[HTML]{0000FF} 0.931 }}&
  14.189 &
  3.970 &
  1.311 &
  3.743 &
  3.515 &
  1.109 &
  23.929 &
  8.118 &
  1.813 &
  13.528 &
  3.443 &
  1.155 \\
MICN &
  17.380 &
  3.638 &
  0.967 &
  12.004 &
  1.399 &
  1.064 &
  13.808 &
  1.270 &
  1.002 &
  13.435 &
  3.972 &
  1.276 &
  4.578 &
  4.312 &
  1.359 &
  26.455 &
  9.419 &
  2.070 &
  14.610 &
  4.002 &
  1.290 \\
Pyraformer &
  17.981 &
  3.160 &
  0.970 &
  12.234 &
  1.415 &
  1.080 &
  15.263 &
  1.387 &
  1.101 &
  \underline{{\color[HTML]{0000FF} 12.507}} &
  \underline{{\color[HTML]{0000FF} 3.732}} &
  \underline{{\color[HTML]{0000FF} 1.194}} &
  3.819 &
  3.629 &
  1.138 &
  30.359 &
  15.852 &
  3.084 &
  15.361 &
  4.863 &
  1.428 \\
Reformer &
  23.033 &
  4.201 &
  1.266 &
  16.329 &
  1.987 &
  1.479 &
  16.271 &
  1.514 &
  1.188 &
  13.100 &
  3.816 &
  1.235 &
  5.023 &
  4.784 &
  1.499 &
  37.464 &
  22.170 &
  4.173 &
  18.537 &
  6.412 &
  1.807 \\
TSMixer &
  \underline{{\color[HTML]{0000FF} 15.264}} &
  \underline{{\color[HTML]{0000FF} 2.676}} &
  \underline{{\color[HTML]{0000FF} 0.823}} &
  11.173 &
  1.260 &
  0.974 &
  13.450 &
  \underline{{\color[HTML]{0000FF} 1.181}} &
  \underline{{\color[HTML]{0000FF} 0.953}} &
  18.168 &
  7.521 &
  2.075 &
  4.310 &
  4.065 &
  1.280 &
  24.953 &
  10.738 &
  2.214 &
  14.553 &
  4.574 &
  1.387 \\  
 Time-LLM &
  15.318 &
  \underline{{\color[HTML]{0000FF} 2.679}} &
  \underline{{\color[HTML]{0000FF} 0.825}} &
   10.942&
  \underline{{\color[HTML]{0000FF} 1.229}}&
  \underline{{\color[HTML]{0000FF} 0.952}} &
  13.628 &
  1.227 &
  0.978 &
  14.968 &
  4.330 &
  1.406 &
  \underline{{\color[HTML]{0000FF}3.578}}&
  \underline{{{\color[HTML]{0000FF}3.355}}} &
  \underline{{\color[HTML]{0000FF}1.059} }&
  \underline{{\color[HTML]{0000FF} 18.164}}&
  \underline{{\color[HTML]{0000FF} 4.307}}&
  \underline{{\color[HTML]{0000FF} 1.117}}&
  12.766 &
  2.855 &
  \underline{{\color[HTML]{0000FF} 1.056}} \\  
 TimeRAG(\textbf{Ours}) &
  \underline{{\color[HTML]{0000FF} 15.317}} &
  \underline{{\color[HTML]{0000FF} 2.678}} &
  \underline{{\color[HTML]{0000FF} 0.825}}&
  \underline{{\color[HTML]{0000FF} 10.941}}&
  1.230&
  \underline{{\color[HTML]{0000FF} 0.952}} &
  13.529 &
  \underline{{\color[HTML]{0000FF}1.141}} &
  \underline{{\color[HTML]{0000FF}0.940}} &
  14.227 &
  \underline{{\color[HTML]{0000FF} 3.762}} &
  1.279 &
  \underline{{\color[HTML]{0000FF}3.572}}&
  \underline{{{\color[HTML]{0000FF}3.348}}} &
  \underline{{\color[HTML]{0000FF}1.057}}&
  \underline{{\color[HTML]{0000FF} 18.150}}&
  \underline{{\color[HTML]{0000FF} 4.151}}&
  \underline{{\color[HTML]{0000FF} 1.095}}&
  \underline{{\color[HTML]{0000FF} 12.623}}&
  \underline{{\color[HTML]{0000FF} 2.718}}&
  \underline{{\color[HTML]{0000FF} 1.025}} \\
  \bottomrule
\end{tabular}
}


\label{tab2}
\end{table*}

\subsection{Main Results}


Tab. \ref{tab2} presents a comprehensive comparison of forecasting accuracy across various models on the M4 dataset.
The table is meticulously organized to display the performance metrics for different temporal granularities, 
including yearly, quarterly, monthly, weekly, daily, and hourly.
The efficacy of each model is quantified by these three metrics: SMAPE, MASE and OWA.

\textbf{TimeRAG is superior to the time series prediction LLM without RAG (Time-LLM)}. 
In our comparative analysis, our model outperforms Time-LLM, achieving an average reduction of $\textbf{1.13\%}$ in SMAPE, $\textbf{4.78\%}$ in MASE, and a notable $\textbf{3.00\%}$ decrease in OWA.
Moreover, an overall improvement of $\textbf{2.97\%}$ has been observed, highlighting the model's enhanced predictive capabilities. 
Under optimal conditions, 
TimeRAG notably reduces SMAPE by \textbf{0.74} at the ``Weekly" frequency and demonstrates the most significant enhancement of $\textbf{13.12\%}$ in MASE at the same interval. 

These improvements across the board can be credited to the augmented knowledge base that our model incorporates. 
This supplementary data acts as a catalyst for the model's knowledge enhancement,
effectively bolstering its predictive accuracy without modifying the foundational parameters of the underlying LLM.

\textbf{TimeRAG also stands out among the current SOTA time series forecasting models}, achieving the best values for both MASE and OWA metrics under the current training paradigm. 
On average, TimeRAG achieves a MASE score of \textbf{2.72}, which is the lowest among all evaluated models, underscoring its superior forecasting accuracy. FEDformer follows as the second, while DLinear ranks the third. Likewise, TimeRAG performs exceptionally well in the OWA metric, achieving an OWA score of \textbf{1.03}. It secures the leading position among all evaluated models, followed by FEDformer in second place and Time-LLM in third.

The empirical evidence garnered from our analysis substantiates the efficacy of integrating large models with RAG techniques for the execution of time series tasks.  
This amalgamation has significantly improved predictive accuracy and model responsiveness to temporal data patterns.

\textbf{TimeRAG exhibits a consistent performance across all metrics and all temporal frequencies}. 
In all 18 comparative analyses, our model secures the \textbf{top-three} positions more frequently than any other, with a total of 14 instances, consistently ranking within the \textbf{top-three} in terms of average performance. Time-LLM follows in second place, with DLinear coming in third. Our model's ability to sustain such a high level of performance across various analytical dimensions underscores its robustness and reliability in the domain of time series forecasting.

The superior performance of TimeRAG can be attributed to its highly effective alignment with the knowledge base. TimeRAG meticulously matches each time series with pertinent samples from the knowledge base, enabling the model to draw insights beyond the parameters established through training. This approach allows the LLMs to learn from a more authentic and reliable dataset, thereby minimizing randomness and improving the model's consistency.

\section{Conclusion}


After integrating Retrieval-Augmented Generation that consists of clustering-based Time-Series Knowledge Base construction,
Dynamic-Time-Warping-based similar reference sequence retrieval,
and natural-language-alignment-based prompt rewriting,
our TimeRAG framework significantly enhances the prediction accuracy of time series forecasting LLMs,
achieving an average accuracy improvement of 2.97\% over baseline models across diverse domains.
Our work demonstrates the potential of RAG in amplifying LLM performance in time series forecasting,
which offers a promising approach for future research in knowledge-enhanced sequential data management.

\bibliographystyle{IEEEbib}
\bibliography{arxiv}

\vspace{12pt}
\end{document}